\documentclass[times, review, 10pt]{elsarticle} 

\usepackage{graphicx}%
\usepackage{multirow}%
\usepackage{amsmath,amssymb,amsfonts}%
\usepackage{amsthm}%
\usepackage{mathrsfs}%
\usepackage[title]{appendix}%
\usepackage{xcolor}%
\usepackage{textcomp}%
\usepackage{manyfoot}%
\usepackage{booktabs}%
\usepackage{algorithm}%
\usepackage{algorithmicx}%
\usepackage{algpseudocode}%
\usepackage{listings}%
\usepackage{tabularx}
\usepackage{array}
\usepackage{makecell}
\usepackage{hyperref}
\usepackage{subcaption}
\usepackage{cleveref}
\usepackage{longtable}
\usepackage{xltabular}
\usepackage{multirow}
\usepackage{array}
\usepackage{threeparttable}
\usepackage{makecell}
\usepackage{refcount}

\newcommand{\repomark}{%
  \footnotemark[\getrefnumber{fn:repo}]%
}

\begin{document}

\newcommand{\eduardo}[1] {\textcolor{blue}{#1}}
\newcommand{\furtado}[1] {\textcolor{red}{#1}}
\newcommand{\alneu}[1] {\textcolor{green}{#1}}
\newcommand{\solange}[1] {\textcolor{yellow}{#1}}
\newcommand{\willam}[1] {\textcolor{black}{#1}}

\begin{frontmatter}

\title{Decomposing images in a PRISM: Single-Source Multi-View Framework for Image Representation}

\author[icmc]{Eduardo Campos\corref{cor1}}
\ead{educampos@usp.br}

\author[icmc]{Willam Cardoso}
\author[icmc]{Diego Furtado}
\author[icmc]{Solange Oliveira}
\author[icmc]{Alneu Lopes}

\cortext[cor1]{Corresponding author}

\affiliation[icmc]{
    organization={Institute of Mathematical and Computer Sciences,
    University of São Paulo},
    city={São Carlos},
    state={São Paulo},
    country={Brazil}
}

\begin{abstract}
Multi-view learning has shown that combining complementary views can improve predictive performance, yet many datasets provide a single raw source of information. How can multiple useful views be constructed from one source? We propose PRISM (Plural Representations of Images via Structural Maps), a framework that transforms an image into representations with distinct inductive biases through explicit structural maps, together with a protocol for measuring what each view contributes. We evaluate PRISM-HD, a hand-designed instantiation of PRISM in which the constructed views are summarized through representation-specific descriptors, on three artistic and four non-artistic image datasets. Analyses at a fixed feature budget show that the gains of the structural views are not explained by feature dimensionality alone. PRISM-HD by itself outperforms WND-CHARM on all seven datasets. Moreover, PRISM-HD and Fisher Vector provide complementary predictive signals: their equal-weight combination improves over the better individual representation by 5--8 percentage points on the artistic datasets. When combined with fine-tuned deep networks through a fixed parameter-free fusion rule, the PRISM-based representation improves performance on four of seven datasets and leaves it statistically unchanged on the remaining three, indicating that the constructed views provide predictive signals that are complementary to learned deep representations.
\end{abstract}


\end{frontmatter}

\section{Introduction}
\label{Intr}

Multi-view learning has shown strong results across a wide range of tasks by combining complementary representations of the same phenomenon
\cite{yu2025review}. Many multi-view methods, however, assume that multiple views are naturally available, such as different sensors, modalities, or acquisition processes \cite{sun2013survey,yu2025review}. Another setting arises when only a single raw source is available, and multiple views must instead be constructed from that source. Although several works apply structural maps that transform one data structure into another, such as images into graphs, sequences, or spectral representations \cite{liu2023towards,lin2025mff,gony2026dual}, these transformations are usually studied in isolation. To the best of our knowledge, no one has comprehensively investigated the extent to which systematically constructing heterogeneous views from a single source yields general and consistent predictive gains.

Motivated by this perspective, we propose PRISM (Plural Representations of Images via Structural Maps), a single-source multi-view framework that applies a set of explicit structural maps to each image, producing heterogeneous representations that include image-based, relational, sequential, spectral, and linguistic views. Except for the linguistic view, these maps create no information that is not already present in the image; what they change is its form. This matters because a learner can only exploit information in the form in which it receives it: regularities that are present in the image but entangled in the pixel grid are effectively unavailable to a model of limited capacity or trained on limited data \cite{xu2020theory}. By making such regularities explicit, the maps shift part of the inductive bias from the learner to the representation, offering a route to useful representations that does not depend on external pretraining. PRISM therefore treats its views not as independent information sources but as reorganizations of the same visual signal under different representational biases, and PRISM-HD evaluated in this paper deliberately summarizes each one with simple descriptors, so as to isolate the contribution of representational diversification itself rather than that of sophisticated cross-view deep learning.

Paintings provide a particularly suitable initial setting for studying this question because their visual structure exhibits several properties that can be naturally exposed through alternative representations. Previous studies have reported approximately isotropic spatial statistics in paintings \cite{koch20101}, as well as spectral properties resembling those observed in natural scenes \cite{graham2010statistical}. Related work has also linked the correlation structure of images to that of one-dimensional series derived from them \cite{silva2021time}. More generally, transformations from images to alternative mathematical domains, such as graphs, sequences, and frequency representations, may discard part of the original signal while making particular structural regularities more explicit to subsequent analysis \cite{gotsman1996metric}.

Our experiments on artistic datasets show that this diversification can substantially change the effectiveness of hand-designed representations. Both classical image descriptors and the homogeneous transformed-image WND-CHARM \cite{ORLOV20081684} are consistently outperformed by the heterogeneous PRISM-HD representation. On two of the three artistic benchmarks, PRISM-HD fused with fisher vector \cite{perronnin2010improving} matches or exceeds the frozen ResNet-50 baseline, while remaining competitive on the remaining dataset. Furthermore, PRISM-HD features provide additional gains when combined with fine-tuned deep representations, indicating that the constructed views capture predictive structure that is complementary to deep learning.

Finally, we evaluate whether these gains are specific to paintings or reflect a more general property of single-source multi-view representation. Experiments on non-artistic image datasets show that the gap to pretrained deep representations varies considerably with the transferability of the pretrained model and the characteristics of the target task. Nevertheless, across these datasets, multi-view aggregation consistently improves over the corresponding single-view hand-designed representation. These results suggest that the value of PRISM-HD lies not only in painting analysis, but more broadly in the idea that a single observation can support multiple predicatively useful representations.

Our main contributions are as follows:

\begin{itemize}

    \item We investigate whether multi-view learning remains beneficial when multiple views are not naturally available, but are instead constructed    from a single underlying data source.
    
    \item We study redundancy and complementarity in the single-source multi-view setting. In particular, we analyse whether different representation families induce distinct learned geometries and how their    predictive contributions overlap when combined.

    \item We examine whether the gains obtained through representational diversification can be explained merely by an increased number of features. By controlling the feature budget, we show that heterogeneous multi-view representations retain an advantage over equally sized single-view feature sets.

    \item We introduce PRISM, a framework for constructing heterogeneous views from a single image through explicit encoding maps. PRISM is designed as a complementary representation framework rather than a replacement for deep learning: its views can be used independently with simple downstream learners or combined with learned deep representations.

\end{itemize}

The remainder of this paper is organized as follows. 
Section \ref{RelWork} reviews the related work. 
Section \ref{PropMethod} introduces the components of PRISM, while Section \ref{Ablation} investigates their integration and complementarity. 
Section \ref{Methodology} describes the experimental methodology, and Section \ref{Results} presents the results. 
Finally, Section \ref{conclusion} concludes the paper.

\section{Related Work}
\label{RelWork}

PRISM sits at the intersection of three lines of work, each of which addresses only part of the problem studied here. Multi-view learning shows that combining complementary views improves prediction, but it usually assumes views given by distinct sources; when views are constructed from a single source, they are typically homogeneous, such as reshaped feature vectors, augmentations of the same image, or banks of image-to-image transforms. Encodings that change the representational domain of an image, such as graphs, sequences, spectra, or text, have been used productively, but few at a time. What is missing is a systematic construction of structurally distinct views from a single image, together with a measurement of how much each contributes beyond the others. Section 2.1 reviews artistic image analysis; Sections 2.2 and 2.3 discuss multi-view learning in general and for paintings; and Section 2.4 covers vision–language models and the construction of the textual view

\subsection{Art classification}
\label{Art}

Rathi et al.~\cite{RATHI2025105626} organize painting-classification methods
into three broad categories: hand-designed feature engineering, conventional
machine learning over extracted descriptors, and end-to-end deep learning.
Early approaches relied on manually specified visual cues, including facial,
color, shape, and brush-stroke features~\cite{711107}, as well as transformed
representations such as Fourier and Chebyshev domains combined with standard
image descriptors~\cite{10.1145/1670671.1670672}. Subsequent work paired such
features with more expressive classifiers, including multilayer perceptrons
and other conventional models~\cite{vandenHerik2000,5967323}. More recently,
deep architectures have become the dominant paradigm, with convolutional
networks and pretrained representations achieving strong results across genre,
medium, artist, and style recognition tasks~\cite{chen2024utilization,
yang2020classification,banerji2016}.

Taken together, this trajectory suggests that improvements in painting
classification have largely been pursued through increasingly expressive
learners. An open question, however, is whether the apparent ceiling of
hand-crafted features reflects an intrinsic limitation of the descriptors
themselves, or partly the representation from which they are extracted. In
this work, paintings therefore serve as an initial setting for investigating
whether multiple heterogeneous views constructed from the same image can
reorganize the available information into more useful representations
to simple downstream descriptors.

\subsection{Multi-view learning}
\label{Multiview}

Multi-view learning involves integrating different aspects of the same phenomenon \cite{yu2025review}. Many areas have seen performance improvements with multi-view approaches, such as facial recognition \cite{7530823}, seizure detection \cite{yuan2018multi}, and vehicle identification \cite{jin2021model}. Recent issues involve view integration, scalability and domain generalisation \cite{yu2025review}.

 In the co-training work, Blum and Mitchell \cite{blum1998combining} exploit two naturally available views of a web page: the words it contains and the anchor text of pages linking to it. Co-training assumes that each view is sufficient for the task and that the views are conditionally independent given the label, so that their combination improves learning. Wang et al. \cite{wang2011novel} extended this perspective to originally single-view data: rather than relying on naturally available sources, they construct multiple matrix representations by reshaping the same feature vector and learn jointly over the resulting views, showing that multi-view learning remains applicable when views must be constructed from a single source.

Other approaches construct views by partitioning an existing feature set, randomly or by optimization \cite{,sun2013survey,ho1998random,nigam2000analyzing}; like Wang et al., they redistribute a fixed representation rather than changing its structure. The most similar line of research applies transformations that change representation bias. Orlov et al. \cite{ORLOV20081684} use image-to-image transformations such as wavelet and Chebyshev to create a general-purpose multi-view image descriptor. Although the idea is an antecedent of this work, the evaluation on homogeneous image-to-image is a limitation compared to PRISM.

These two settings differ in where view diversity comes from. In co-training, it arises from distinct ways of observing the object; in Wang et al., it is induced by alternative reorganizations of the same vector, which remain within a single representational structure. Unlike classical co-training, PRISM does not assume conditional independence between constructed views. Since internally derived views are transformations of the same underlying observation, their dependence and redundancy are treated as empirical properties and measured explicitly. Self-supervised methods also derive views from a single image \cite{chen2020simple,tian2020contrastive}, but these are homogeneous image-to-image transformations used to learn invariances, that is, to discard what differs between views. PRISM generalizes the constructed-view principle in the opposite direction: it builds structurally distinct views, such as graphs, sequences, spectra, and text, so that each transformation introduces a different inductive bias and exposes different properties of the same data.

\subsection{Multi-view learning for paintings}
\label{MultiviewArt}

Yang et al.~\cite{yang2018historical} incorporate historical context, including temporal, geographical, and art-movement information, as additional views through auxiliary label distributions used during training. Their formulation therefore combines the visual observation with external contextual views, whereas PRISM focuses primarily on constructing multiple representations from the same underlying image. Chu et al. \cite{chu2018image} propose a multi-view representation based on different correlation statistics computed over feature maps extracted by a VGG network. While these statistics provide alternative representations of the same activation space, they remain within a common deep visual feature domain. PRISM differs from this formulation by constructing views in structurally heterogeneous representational domains.

Wang and Feng \cite{wang2025multimodal} propose MSA-Net, a multi-stream
architecture that combines texture, structural, and color representations. Texture is modeled through multi-scale deep correlation features and a Transformer; structural information is obtained from deep visual features, and color is represented through RGB histograms. This formulation is closely related to PRISM in that multiple complementary representations of the same image are explicitly fused. However, its streams characterize different visual properties while remaining within a common visual feature domain.

\subsection{Text as a derived representation}
\label{TextGen}

Language has increasingly been used as an intermediate visual representation.
Contrastive image--text pretraining showed that linguistic supervision can
produce transferable visual representations~\cite{radford2021learning}, while
models such as BLIP-2 enable image-conditioned text generation without
task-specific adaptation~\cite{li2023blip}. Related approaches use language to
improve zero-shot recognition through class descriptions or automatically
generated prompts~\cite{menon2022visual,pratt2023does}. In these formulations,
however, text typically describes the target classes and mediates access to the
label space. PRISM instead generates a description for each image
and treats it as an additional, externally informed view of the observation,
alongside representations derived from other structural domains.

Taken together, prior work shows a progression from richer classifiers to richer representations of artistic images. Recent multi-view approaches further demonstrate the value of combining complementary views, but typically rely on external context, homogeneous visual transformations, or task-specific view designs. PRISM builds on this line of work by studying heterogeneous single-source view construction as a general representation problem, spanning relational, spectral, sequential, and textual domains.
\section{Plural Representations of Images via Structural Maps  (PRISM)}
\label{PropMethod}

\begin{figure}[!htbp]
    \centering
    \includegraphics[width=1\linewidth]{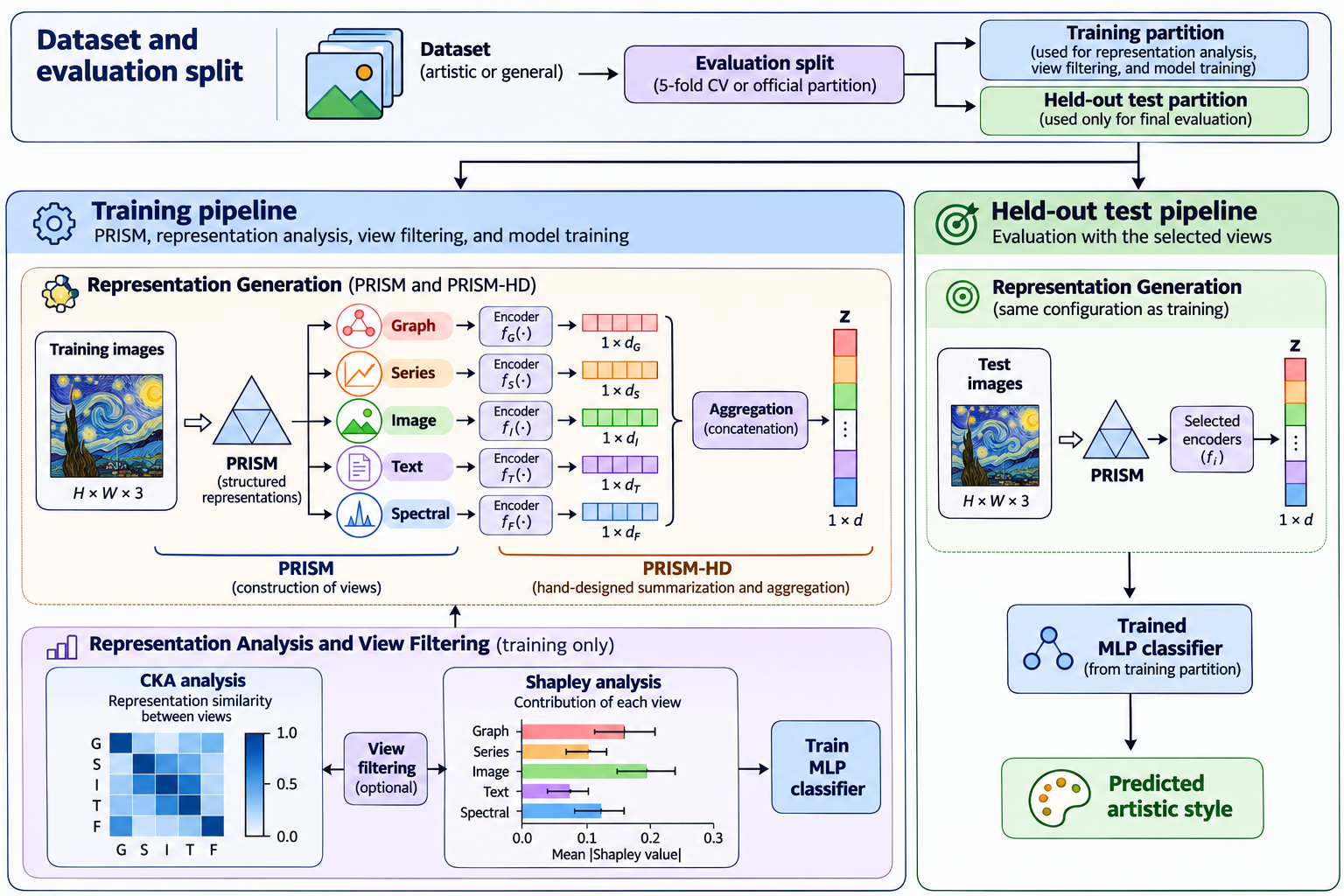}
    \caption{Pipeline summarizing the PRISM framework. The pipeline decomposes an initial image into multiple views and uses hand-designed features to summarize those views in an early-fusion configuration. The workflow diagram was created with assistance from OpenAI's image-generation tool and subsequently reviewed and edited by the authors, The CKA and Shapley plots are schematic and their values are illustrative only.}
    \label{modalitis}
\end{figure}

We introduce PRISM (Plural Representations of Images via Structural Maps), illustrated in Figure \ref{modalitis}. Rather than relying exclusively on the original image representation, PRISM maps each image into multiple derived views, each grounded in a different assumption about how visual information can be organized. These transformations make explicit structural regularities that remain implicit in the original pixel space, allowing even less complex hand-designed descriptors to capture the information.

Starting from the same input image, PRISM derives a set of alternative representations, including relational, sequential, spectral, and linguistic views. For each derived data type, as well as for the original image, an encoder maps the representation into a vector space intended to summarize the information exposed by that view. Using only the training partition, we then estimate representational similarity through CKA and view contribution through Shapley analysis, which are used to identify redundant or weakly contributing representations. After view selection, a branch-based MLP jointly encodes the retained representations into a shared predictive space.

In this study, we instantiate the general framework as PRISM-HD, a version of PRISM in which each view is summarized using hand-designed tabular descriptors before fusion. Table~\ref{tab:encoding_maps} summarizes the structural transformations considered in this work and illustrated in Figure~\ref{usedTranfs}; since most of them are well established, further details are provided in the Supplementary Appendix \repomark. This configuration was deliberately chosen to reduce confounding between gains due to representational diversification and gains introduced by highly expressive view-specific learners, allowing the contribution of the constructed views themselves to be evaluated more directly.

\subsection{Structural Maps}
\label{Encoding}

We characterize a family of structural maps $\sigma_{\epsilon}$, parameterized by $\epsilon$, operating between representation spaces. Let $\mathcal{X}_i$ denote the space associated with view type $i$, and let $v_i^{(n)} \in \mathcal{X}_i$ denote the realization of that view for instance $n$. The corresponding dataset-level realization is

\begin{equation}
V_i =
\left\{v_i^{(n)}\right\}_{n=1}^{N}
\in \mathcal{X}_i^N.
\end{equation}

A structural map may therefore operate either on an individual realization,

\begin{equation}
\sigma_{\epsilon}:
v_i^{(n)} \mapsto v_j^{(m)},
\qquad
v_i^{(n)} \in \mathcal{X}_i,
\quad
v_j^{(m)} \in \mathcal{X}_j.
\label{eq:encoding_map}
\end{equation}

or jointly on a collection of realizations,

\begin{equation}
V_i =
\left\{v_i^{(n)}\right\}_{n=1}^{N},
\qquad
\sigma_{\epsilon}: V_i \mapsto v_j.
\label{eq:global_map}
\end{equation}

We organize these maps according to three orthogonal axes:
information source, cardinality, and domain preservation. The first axis concerns the source of information. An internal map derives its output exclusively from information available within the benchmark, including the native observations and, when applicable, the structure or supervision of the training set. An external map, in contrast, incorporates information acquired outside the benchmark, such as parameters learned from an auxiliary corpus or metadata not provided as part of the task.

For an internal transformation, no information can be created beyond that available to the mapping itself. If a view is obtained solely from an instance $v_i^{n}$, the data-processing inequality gives.

\begin{equation}
I(Y;\sigma_{\epsilon}(v_i^{n}))
\leq
I(Y;v_i^{n}).
\label{eq4}
\end{equation}

More generally, for dataset-level transformations that additionally depend
on the training set $\mathcal{D} {\mathrm{tr}} \in \mathcal{X}_i$,

\begin{equation}
I(Y;\sigma_{\epsilon}(v_i^{n},\mathcal{D}_{\mathrm{tr}}))
\leq
I(Y;v_i^{n},\mathcal{D}_{\mathrm{tr}}).
\end{equation}

Nevertheless, under computational constraints, a transformation may increase the information that is usable by a restricted predictive family \cite{xu2020theory}, by reorganizing existing information into a form that is more accessible to the downstream learner.

The second axis concerns the cardinality of the transformation. An instance-level ($1{:}1$) map independently transforms each instance into one corresponding derived representation. In contrast, a global map ($N{:}1$) encodes an entire dataset to a single instance.

The third axis concerns domain preservation. A
homogeneous map preserves the underlying data structure of its input, even if it changes the domain in which its information is expressed. For example, a Fourier transformation maps a spatial image grid to a frequency-domain grid. A heterogeneous map, in contrast, changes the underlying representational structure, as occurs when an image is
mapped to a graph, a sequence, or a linguistic description. 

A final remark concerns the depth of a derived view. A view obtained directly from the native representation has depth one, whereas successive structural maps increase the compositional depth. For example, a series derived from an image can be written as $\sigma_{\epsilon_1}^{(1)}(v)$, while a graph subsequently constructed from that series is obtained as $\sigma_{\epsilon_2}^{(2)}\!\left(\sigma_{\epsilon_1}^{(1)}(v)\right)$.

These categories have been explored unevenly in the literature. Most of the transformations identified in our review are internal and instance-level, whereas global and heterogeneous mappings have received comparatively less attention. The transformations considered in PRISM are not intended to constitute an exhaustive enumeration of all possible structural maps; rather, they provide a representative set spanning the categories relevant to the scope of this work.

PRISM does not prescribe a particular strategy for encoding or combining the constructed views. In this study, we instantiate the framework as PRISM-HD, in which each representation is summarized through hand-designed tabular descriptors before fusion. More expressive representation-specific encoders and alternative fusion strategies remain compatible with the general framework.

 \begin{figure}[!htbp]
    \centering
    \includegraphics[width=0.8\linewidth]{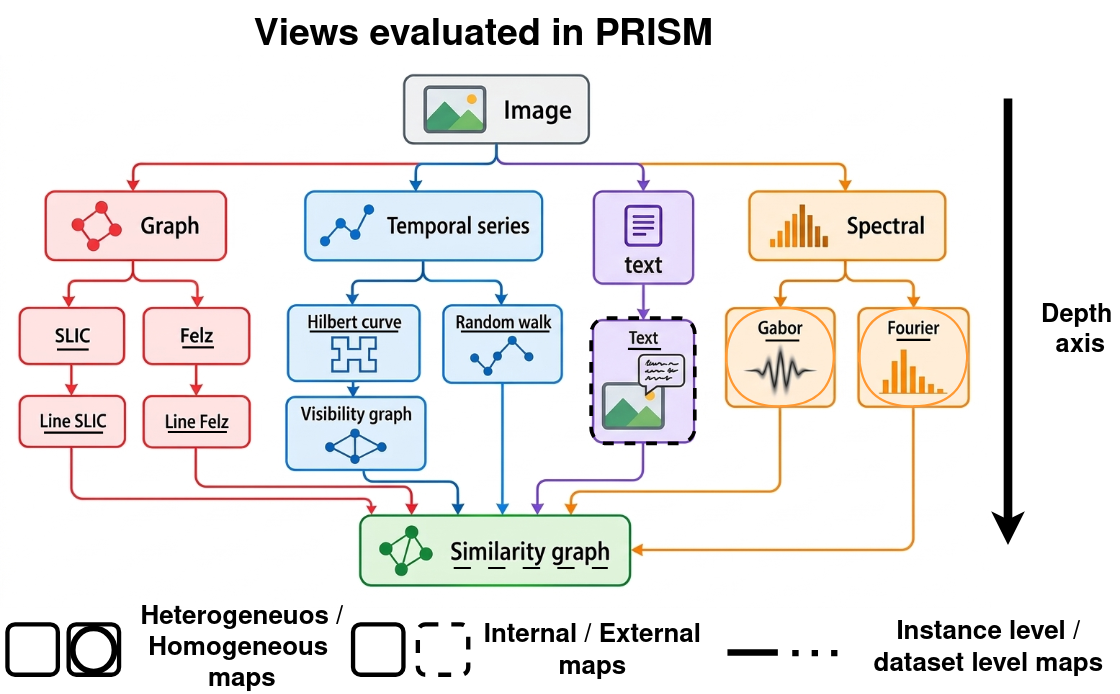}
    \caption{Summarization of the views evaluated in PRISM-HD by our taxonomy.}
    \label{usedTranfs}
\end{figure}

\begin{table}[!htbp]
\centering
\caption{Structural maps used in PRISM-HD.}
\label{tab:encoding_maps}

\renewcommand{\arraystretch}{0.9}
\setlength{\tabcolsep}{6pt}

\begin{tabularx}{\textwidth}{
    >{\raggedright\arraybackslash}p{0.20\textwidth}
    >{\raggedright\arraybackslash}X
}
\toprule
\textbf{Method} & \textbf{Description} \\
\midrule

Image statistics
& Direct descriptors computed from the native pixel grid, capturing texture,
gradients, and intensity organization. \\
SLIC & Region-adjacency graph from superpixels. \\
Felzenszwalb & Region-adjacency graph from graph-based segmentation. \\
Hilbert curve & Locality-preserving 1D pixel ordering. \\
Random walk & Stochastic local pixel traversal. \\
Fourier & Global frequency-domain representation. \\
Gabor & Localized orientation-frequency representation. \\
Line graph & Graph whose nodes represent original edges. \\
HVG & Visibility graph derived from a sequence. \\
Similarity graph & Dataset-level instance similarity graph. \\
Image description & Externally informed linguistic representation. \\
\bottomrule
\end{tabularx}
\end{table}


\subsection{What Constitutes a Representation-Specific Feature?}
\label{what}

A purely ontological definition could classify any scalar derived, directly or indirectly, from an image as an image feature. Although internally consistent, such a definition is of limited use in a multi-view setting. If every deterministic transformation of a pixel grid is reduced to its source, then graph, sequential, and spectral representations become indistinguishable from conventional image descriptors simply because they ultimately originate from the same input.

PRISM therefore distinguishes between feature origin and
feature behavior. Feature origin describes the representation
from which a descriptor is explicitly computed. For example, mean node  degree is a graph-origin feature because it is computed from a graph, even when that graph itself was deterministically constructed from an  image. Feature behavior, in contrast, describes the statistical  neighborhood occupied by a feature with respect to the complete feature  set.

This distinction is important because representations derived from the same image are neither independent nor necessarily redundant. A
descriptor computed directly from pixels may behave similarly to graph, spectral, or sequential descriptors if it captures a regularity that is also made explicit by those transformations. Conversely, features derived from the same representation may exhibit strong mutual redundancy. Thus, the role of a structural map is not to create information that was absent from the original image, but to reorganize  existing information so that particular structural regularities become explicitly accessible to downstream descriptors.

Accordingly, PRISM treats representation labels as descriptions of the computational path by which features are obtained, rather than as claims  of informational independence. Their empirical distinctiveness is assessed separately through the redundancy structure of the complete  feature space. This allows us to examine whether the proposed views  produce genuinely different feature organizations, rather than assuming such complementarity from their construction alone.

\section{Experimental Setup}
\label{Methodology}
This section discusses the experimental settings.

\subsection{Datasets}

The three artistic corpora were selected to differ along dimensions that are relevant to the proposed framework rather than to represent repeated instances of the same classification problem. They vary in label semantics, visual medium, class structure, and image scale. Multitask125 \cite{bianco2019multitask} and OilPainting \cite{chu2018image} use artistic style as the target label, whereas Rijskdraw  \cite{mensink2014rijksmuseum} is organized by historical period, introducing a substantially different semantic criterion. The datasets also differ in medium: Multitask125 and OilPainting contain paintings, while Rijskdraw consists primarily of drawings, where line and texture play a more prominent role than color.

The corpora further span markedly different levels of class granularity and imbalance. Multitask125 contains 125 classes, compared with 17 in OilPainting and 11 in Rijskdraw, with correspondingly different class-frequency distributions. This variation is particularly relevant for representations whose behavior depends on neighborhood density, such as the similarity-graph view. Finally, the datasets differ substantially in both size and image resolution. Together, these variations provide a heterogeneous artistic evaluation setting for assessing the behavior of PRISM-HD under different semantic, structural, and
visual conditions.

The generalization experiment introduces four non-artistic datasets chosen to contrast tasks dominated by visual appearance with those requiring more semantic recognition. NCT-CRC-HE-100K \cite{kather2019predicting} contains 100,000 histopathological patches from colorectal tissue distributed over nine tissue categories. Its classes are largely characterized by recurring morphological and textural patterns distributed throughout the image. MINC-2500 \cite{bell2015material} similarly focuses on material recognition, comprising 23 material categories with 2,500 image patches per class, where surface texture, color, and appearance provide much of the discriminative signal.

In contrast, Food-101  \cite{bossard2014food} contains 101 food categories and 101,000 images, requiring recognition of particular food objects despite substantial variation in presentation and background. Places-LT  \cite{liu2019large} addresses long-tailed scene recognition over 365 categories and 62,500 training images, introducing both stronger semantic requirements and severe class imbalance. Together,
these datasets allow us to examine whether the behavior observed in artistic images extends from predominantly texture- and appearance-driven tasks to object- and scene-centric recognition problems.

\subsection{Baselines and Pretraining Regimes}
Two families of methods were selected for comparison with PRISM-HD. The first
comprises classical hand-crafted descriptors, which provide the closest
methodological comparison to our framework because they do not rely on
external pretraining and encode manually specified assumptions about image
structure. Although domain-specific descriptors can exploit properties
particular to a given application, empirical studies in painting
classification suggest that such specialization often provides incremental
rather than order-of-magnitude improvements over conventional hand-crafted
representations \cite{kinoshita2026painting}. We therefore selected Haralick
features and Histogram of Oriented Gradients (HOG) as representative
baselines due to their long-standing and widespread use in image analysis.

We also include two additional classical representation methods to provide a stronger evaluation. WND-CHARM is a particularly relevant antecedent of PRISM, as it applies multiple image transformations before feature extraction, but remains largely restricted to homogeneous image-to-image mappings~\cite{ORLOV20081684}. Its inclusion therefore helps assess whether extending representation diversity beyond transformed image planes provides additional benefits. Fisher Vector serves a complementary role: although it is not directly related to PRISM's multi-view formulation, it was among the strongest shallow image representations before the widespread dominance of deep convolutional networks~\cite{Chatfield2014ReturnOT}. Together, these baselines provide a more demanding comparison against both transform-based representation diversification and high-performing pre-deep feature encoding.

The second family comprises deep learning models. These methods represent the dominant performance regime in contemporary image classification benchmarks, but typically rely on substantially larger parameter counts and offer more limited feature-level interpretability. We selected ResNet-50 \cite{he2016deep} and ViT-B/16 \cite{dosovitskiy2020image} as representative convolutional and transformer-based baselines, respectively. Rather than adopting painting-specific deep architectures, we use these widely established models because previous work reports only modest differences between generic deep backbones and specialized painting-classification models \cite{wang2025multimodal}. This choice provides a stronger and more transferable reference for evaluating the contribution of PRISM.

To keep the comparison fair, we classify every feature-based condition with the same network, hyperparameters, folds, and seed; only the input blocks change. The classifier is a branch MLP: each representation is standardised (statistics fitted on the training rows only) and passed through its own encoder, Linear($d_i \rightarrow 128) \rightarrow BatchNorm \rightarrow GELU \rightarrow Dropout(0.25)$; the encoder outputs are concatenated and fed to a shared head of $512 \rightarrow 256 \rightarrow 128 \rightarrow C$, with BatchNorm, GELU and Dropout(0.35) between layers. Single-representation baselines — Haralick, HOG, and the frozen backbones — enter as a single branch, so the branch structure follows the number of views but nothing else differs. 

Training uses AdamW (lr $3·10^{-4}$, weight decay $10^{-4}$), a cosine schedule over the epoch budget, batch size 4096, cross-entropy with label smoothing 0.05 and inverse-frequency class weights, for at most 150 epochs; the reported epoch is selected on a 10\% stratified validation split held out from the training folds, with patience 15. For the artistic datasets without an official train/test partition, results are obtained using 5-fold stratified cross-validation with five random seeds. For the remaining datasets, we follow the official benchmark evaluation protocol and report results over five random seeds. Since the datasets lie in a wide range of class imbalance, all classifiers were tuned to maximize balanced accuracy; normal accuracy and F1-score are displayed as complementary values in the Supplementary Appendix \repomark, since all metrics behave similarly. 

ResNet-50 and ViT-B/16 were initialized from supervised ImageNet-1k weights and fine-tuned end-to-end at \(224\times224\) resolution. Training used RandomResizedCrop, horizontal flipping, and RandAugment, while evaluation used resize and center crop. Class imbalance was addressed through inverse-frequency weighted cross-entropy, and checkpoints were selected according to validation balanced accuracy. ResNet-50 was optimized with SGD and momentum, whereas ViT-B/16 used AdamW with label smoothing. Results are averaged over five random seeds; datasets without predefined splits additionally use five-fold stratified cross-validation. Full optimization parameters and dataset-specific stopping policies are reported in the Supplementary Appendix, the full script is reported under the Github  Repository\footnote{\label{fn:repo}%
\href{https://github.com/Dudu-Campos/Paintings}
{Repository containing experiment setup and hyperparameters.}}

\subsection{Ablation Study}

To characterize both the individual views and their role within the complete framework, we conduct an extensive ablation study. Section~\ref{Ablation} first evaluates the predictive performance of each view in isolation across multiple classifier families. We then analyze the complementarity and redundancy among the resulting representations. Finally, the views are combined into the full PRISM model, and we examine whether the observed gains persist under a fixed feature budget. 

Additional robustness and diagnostic analyses, including textual-channel leakage, computational cost, and sensitivity to representation parameters, are reported in the Supplementary Appendix\repomark. For that,  the experiments were run on an Intel I9-14k with 125 GB RAM and an NVIDIA RTX A6000, 48 GB.
\section{Ablation}
\label{Ablation}

This section discusses the integration of the derived representations into our final model. The analysis shows that the signal captured by each family is complementary and redundancy occurs within the same family. This characteristic leads to fusion gains. All experiments were evaluated on the Multitask125 dataset.

\subsection{Views performance}

\begin{figure}[!htbp]
    \centering
    \includegraphics[width=0.6\linewidth]{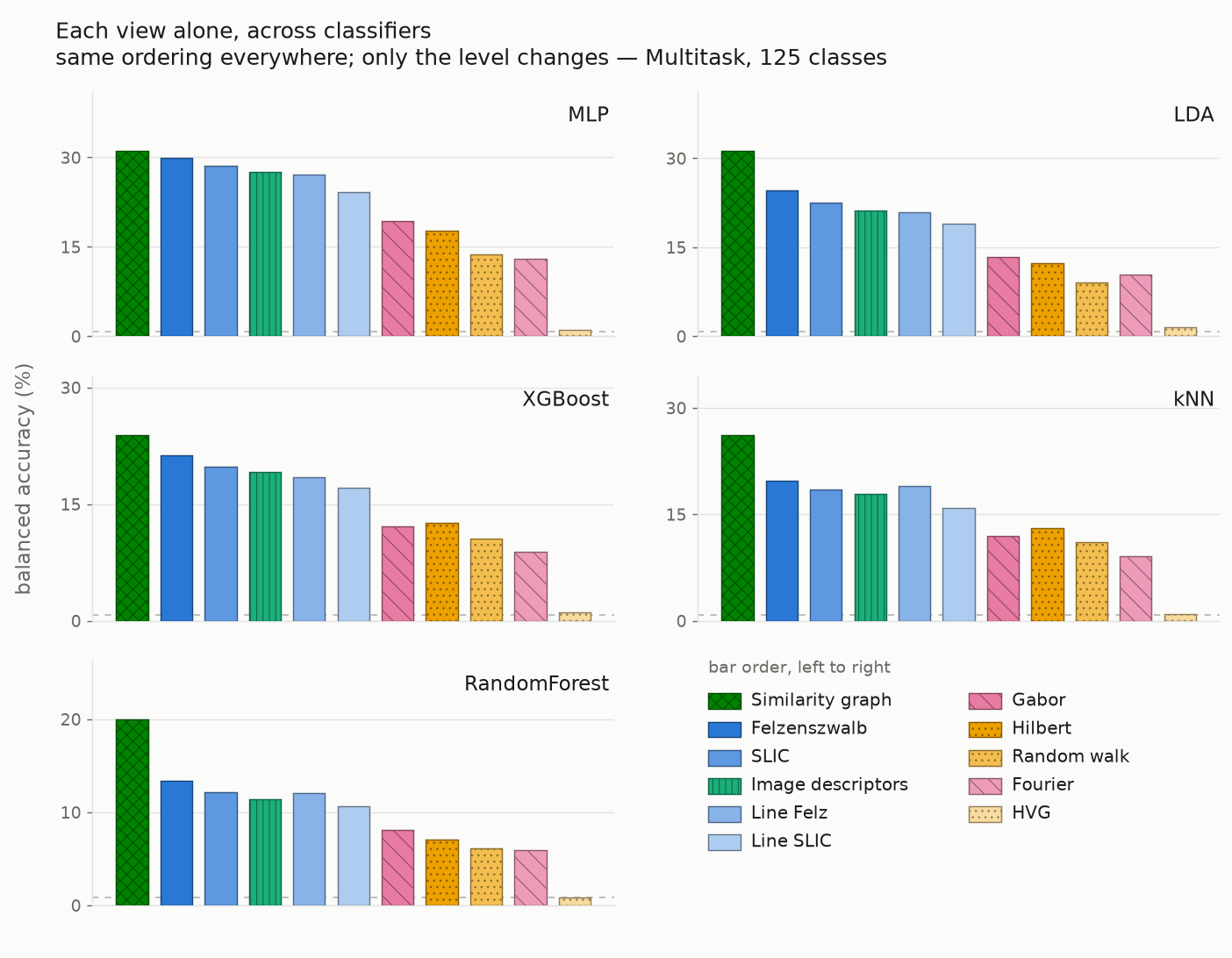}
    \caption{Single-view performance across classifiers remains highly stable.}
    \label{singularPerformance}
\end{figure}

Figure \ref{singularPerformance} reports the individual balanced accuracy of each view across the classifier set \{\text{RF}, \text{MLP}, \text{XGBOOST}, \text{LDA}, \text{KNN}\}. Although the absolute performance of a given view varies with the downstream classifier, the relative ranking of the views remains highly consistent across models. The purpose of this analysis is therefore not to establish a performance hierarchy among the views, since both the number and the specific choice of descriptors within each representation are design-dependent, but rather to assess the stability of their relative behavior across different classifier families. Given this property, the following analysis will evaluate only the more complex and simpler models \{MLP, LDA\}.

The analysis shows that every view performs above the random baseline across all classifier families. A particularly notable result is the strong performance of the similarity-graph representation under LDA, which approaches the MLP ceiling despite relying on a substantially simpler classifier. However, unlike the other views, the similarity graph is not constructed independently: its edge weights are defined from similarities computed over the remaining feature representations. Its performance therefore reflects, at least in part, information already encoded by those features, rather than the contribution of an entirely self-contained representation.

\subsection{Redundancy among families}

Although the individual views achieve substantial predictive performance, part of their signal is expected to overlap; consequently, the benefit of combining them cannot be interpreted as an intersection-free sum of their individual contributions. 

\begin{figure}[!htbp]
    \centering
    \begin{subfigure}[b]{0.48\textwidth}
        \centering
        \includegraphics[width=\textwidth]{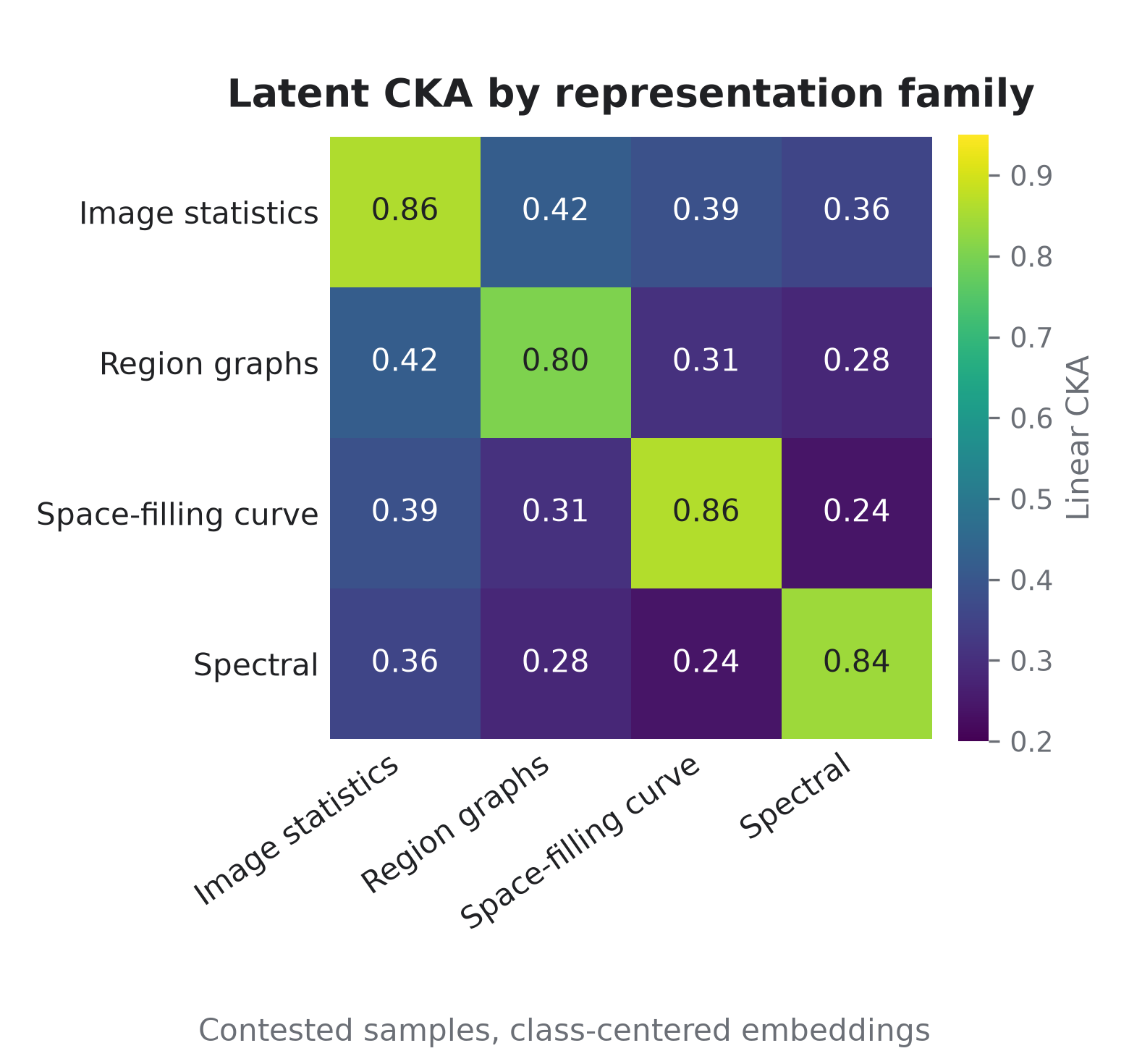}
        \caption{MLP CKA across families.}
        \label{Agreement}
    \end{subfigure}
    \hfill
    \begin{subfigure}[b]{0.48\textwidth}
        \centering
        \includegraphics[width=\textwidth]{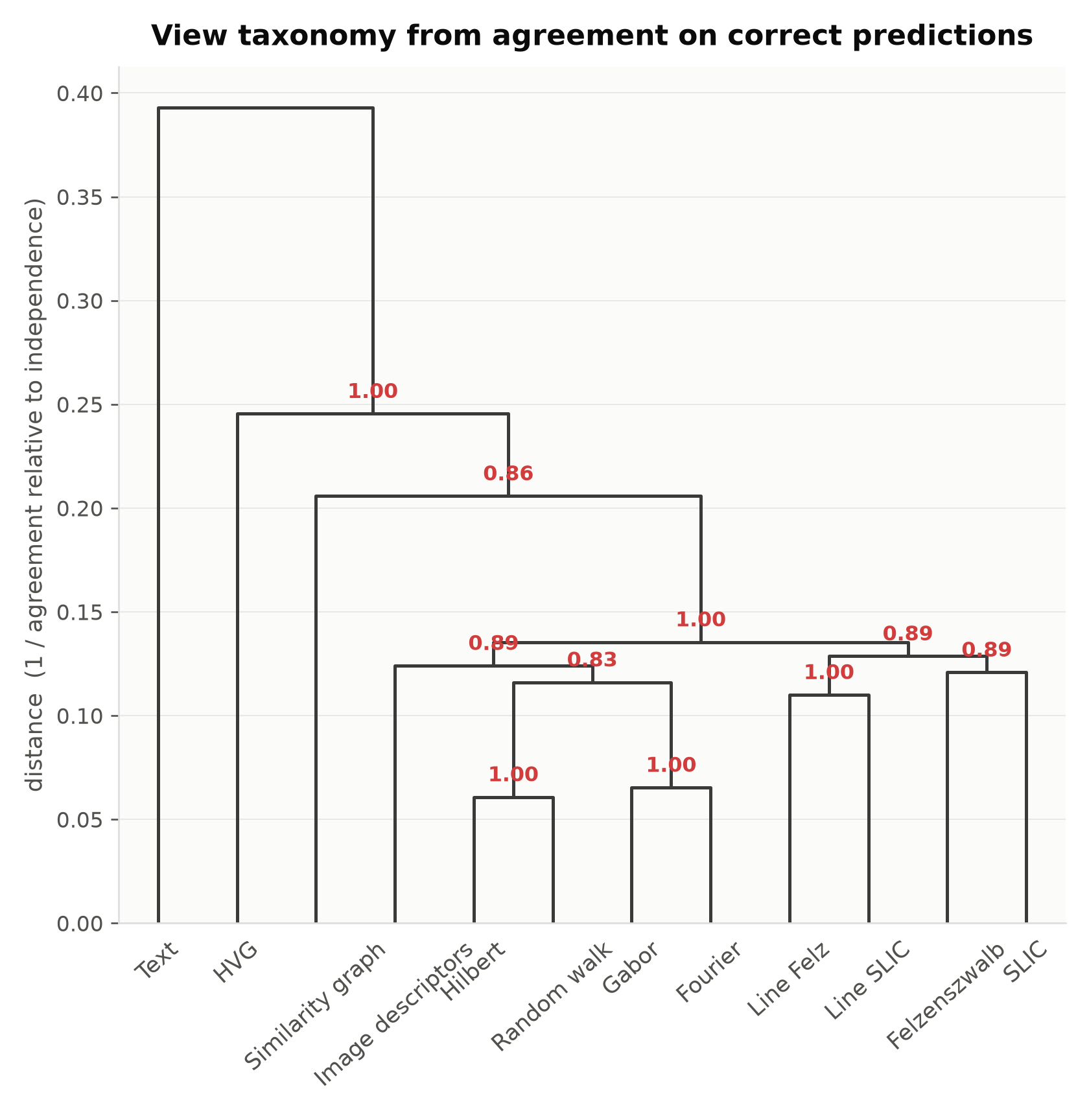}
        \caption{views family dendrogram.}
        \label{dendogram}
    \end{subfigure}
    \caption{Both the representation space and the prediction agreement depend on the view family. }
    \label{fig:imagens_lado_a_lado}
\end{figure}

Figure~\ref{Agreement} compares the latent representation spaces learned by
the same MLP architecture using linear CKA. Across independent training seeds, models trained on representations from the same family produce highly similar latent spaces, whereas the similarity between different families is substantially lower. This indicates that the observed differences are not merely a consequence of random initialization, but are systematically induced by the underlying representation family. These results suggest that the encoding maps lead the same downstream learner to organize the same set of images according to different latent geometries. The views should therefore be regarded as partially overlapping rather than independent, while still exposing sufficiently distinct representational structures to motivate their joint use.

This representational structure is also reflected in the prediction space. Figure~\ref{dendogram} shows a dendrogram constructed from pairwise agreement on correct predictions, measured relative to the agreement expected under independence. The resulting organization largely recovers the expected view taxonomy: region-graph, spectral, and space-filling representations form coherent groups, while image descriptors remain comparatively close to the internally derived views.

Text and the Similarity Graph are substantially more dissimilar from the
remaining representations, which is consistent with their distinct information sources: the former relies on externally pretrained models, whereas the latter incorporates dataset-level relational information. HVG also appears as an outlier; however, given its low standalone discriminative performance, this separation should not be interpreted as evidence of useful complementarity.

\begin{figure}[!htbp]
    \centering
    \includegraphics[width=0.5\linewidth]{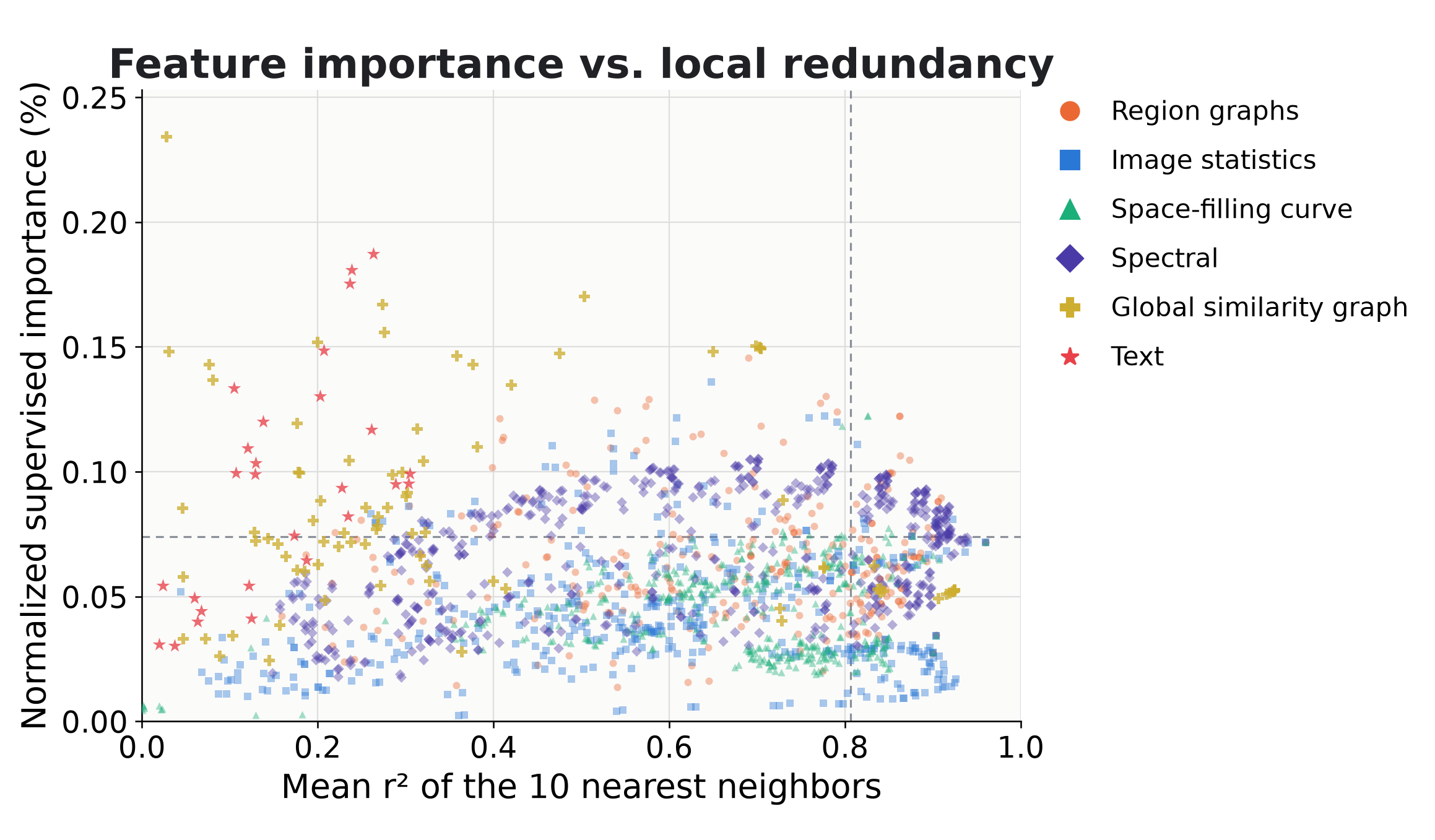}
    \caption{Feature importance and redundancy scatter.}
    \label{featureImportance}
\end{figure}

Figure \ref{featureImportance} scatters local redundancy against feature importance, as measured by an LDA classifier. The Global similarity graph and Text channels occupy the high-importance, low-redundancy region. This finding is congruent with our taxonomy: these are the two views that extract out-of-image information — one via similarity among instances (the global similarity graph), and the other via a pretrained model (Text). The image also shows the heterogeneous aspect of image descriptors; the majority lie in low importance and high redundancy, and some are scattered all over the plot.

\subsection{Fusion performance}

\begin{figure}[!htbp]
    \centering
    \includegraphics[width=0.5\linewidth]{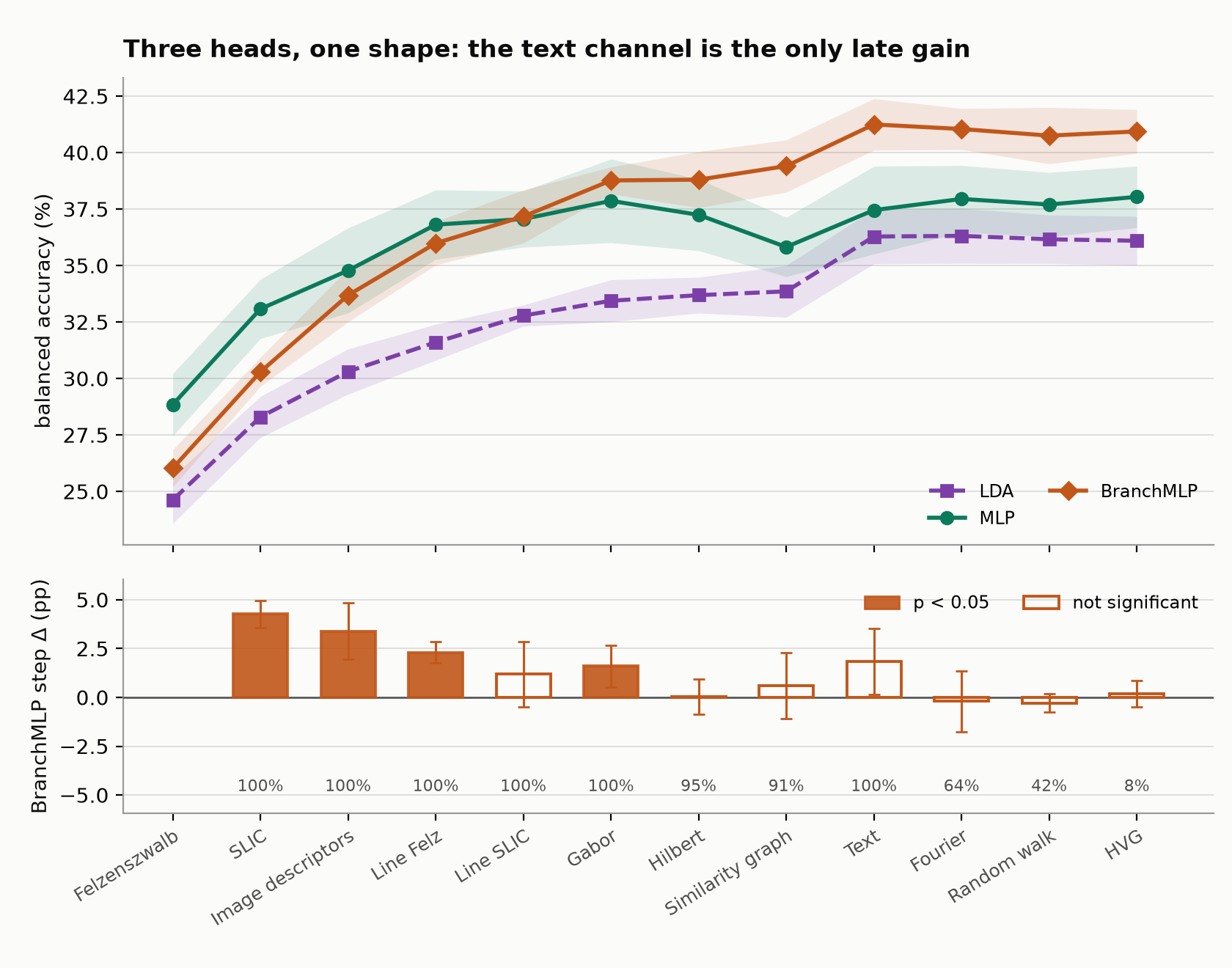}
    \caption{Cumulative ell performance across classifiers.}
    \label{cumShap}
\end{figure}

Figure \ref{cumShap} presents the cumulative contribution curves obtained
with LDA, a standard MLP, and an MLP with independent per-view encoding
branches. Across all three models, the final three views (Fourier, Random
Walk, and HVG) provide no measurable benefit when added to the preceding
representations. In the standard MLP, the Similarity Graph and Hilbert views
also lead to a decrease in performance. This degradation disappears when
each view is processed by a separate encoder before fusion, under which
their marginal contributions become statistically non-significant rather
than negative. 

These results suggest that part of the observed degradation
may arise from interference between heterogeneous views when they compete
for a shared representation space, rather than from the views being
intrinsically detrimental. Those findings and the Shapley analyses shown in the Supplementary Appendix \repomark lead us to remove the latter two from the final PRISM-HD. The Similarity Graph was also removed for a different reason: it is computationally expensive and provides only a small amount of additional signal not already captured by the remaining views.

\subsection{Size confound control}

A potential confound in the multi-view evaluation is feature dimensionality. Since fusion increases the number of available features, part of the observed gain could in principle arise simply from exposing the classifier to a larger input space, particularly in data-rich settings such as Multitask125.

\begin{figure}[!htbp]
    \centering
    \includegraphics[width=1.0\linewidth]{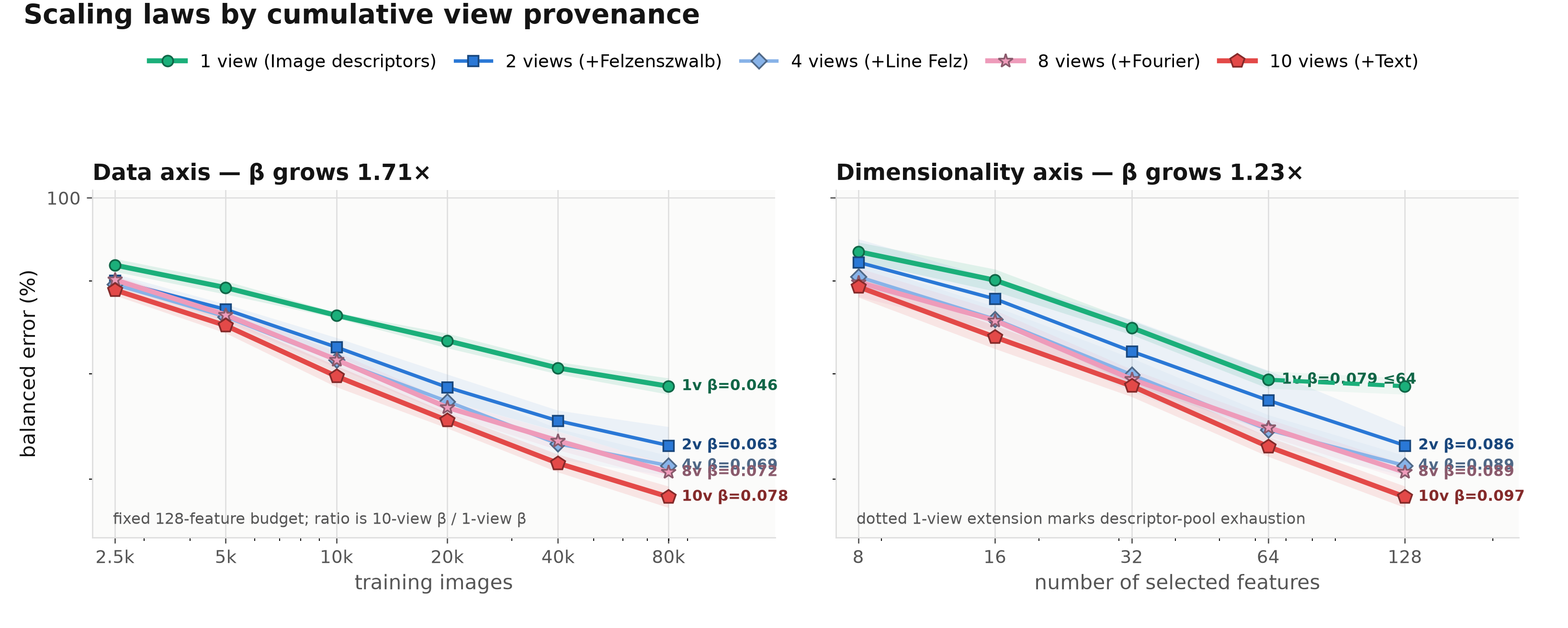}
    \caption{LDA performance changing the number of training data and the number of features.}
    \label{scalingLaw}
\end{figure}

Figure~\ref{scalingLaw} quantifies this effect. On the right, increasing the number of input features reduces classification error for both the image-descriptor baseline and PRISM. Although the estimated dimensional scaling slopes differ significantly ($p=0.036$), the difference in magnitude is modest. 

A substantially clearer separation emerges along the data axis. When the feature dimensionality is fixed at 128, the LDA error decreases considerably faster in the multi-view setting ($p<0.0001$). This suggests that the advantage of PRISM is more strongly associated with how heterogeneous representations exploit additional training data than with simply increasing the dimensionality of the feature space. Although, as discussed in Sec.~\ref{what}, this type of analysis is inherently dependent on how a view is defined and on the specific features extracted from each representation. A second limitation is classifier dependence, as shown in Figure \ref{cumShap}; the utility of a view is not only determined by the dataset and the task, but also by the ability of the classifier.

\section{Results}
\label{Results}

This section presents the comparative evaluation of PRISM-HD against classical hand-crafted methods and pretrained deep-learning models across both artistic and external datasets. A potential concern is that Multitask125 is used both for exploratory analyses and as part of the final evaluation, and therefore its results should not be interpreted as fully independent of the framework development process.

The remaining six datasets provide a stricter assessment of transfer to distributions not involved in the exploratory design of PRISM-HD. Nevertheless, the analyses in Supplementary Appendix .\repomark. indicate that the conclusions drawn from Multitask125 are not strongly dependent on narrowly optimized choices: perturbations of individual transformation parameters are largely absorbed under multi-view fusion, including configurations substantially different from the selected $\theta^{\ast}$.

Similarly, the view-selection analysis in Figure~\ref{cumShap} shows that even views excluded from the final configuration tend to have only a small negative effect when retained. Together, these results reduce, although do not eliminate, the possibility that fine-grained configuration choices drive the reported performance specific to Multitask125.

\subsection{Hand-craft representation evaluation}    

Table~\ref{main} presents the comparative evaluation of PRISM-HD and other hand-designed representations. As expected, Haralick descriptors and HOG provide relatively weak baselines, performing approximately 20 percentage points below Fisher Vector on several datasets. WND-CHARM, the homogeneous predecessor most closely related to PRISM-HD, consistently narrows this gap across all seven datasets by providing a more diverse transformed representation, although it generally remains below the strongest methods.

\begin{table}[!htbp]
\centering
\caption{
Balanced accuracy (\%) of classical and PRISM-based representations.
$\oplus$ denotes fixed equal-weight posterior fusion, and $+t$ denotes
the externally informed textual view.
}
\label{main}

\small
\renewcommand{\arraystretch}{1.08}

\begin{tabular}{lccc}
\toprule
\textbf{Model}
& \textbf{OilPainting}
& \textbf{Multitask125}
& \textbf{Rijksdraw} \\
\midrule

Haralick
& $21.0 \pm 1.5$
& $16.1 \pm 1.2$
& $33.4 \pm 0.8$ \\

HOG
& $20.8 \pm 0.8$
& $13.2 \pm 0.7$
& $30.6 \pm 0.7$ \\

WND-CHARM
& $35.0 \pm 0.8$
& $31.4 \pm 0.9$
& $47.1 \pm 1.0$ \\

Fisher Vector
& $46.8 \pm 0.8$
& $40.1 \pm 1.1$
& $54.4 \pm 0.3$ \\

\midrule

PRISM-HD
& $43.4 \pm 1.4$
& $33.1 \pm 0.7$
& $55.8 \pm 0.8$ \\

\makecell[l]{PRISM-HD $\oplus$\\ Fisher Vector}
& $52.3 \pm 0.9$
& $47.8 \pm 1.0$
& $61.4 \pm 0.3$ \\

PRISM-HD$_{+t}$
& $47.7 \pm 1.1$
& $40.2 \pm 1.2$
& $57.3 \pm 0.5$ \\

\makecell[l]{\textbf{PRISM-HD$_{+t}$ $\oplus$}\\
\textbf{Fisher Vector}}
& $\mathbf{54.3 \pm 0.8}$
& $\mathbf{49.4 \pm 0.9}$
& $\mathbf{61.6 \pm 0.6}$ \\

\bottomrule
\end{tabular}

\vspace{0.6em}

\begin{tabular}{lcccc}
\toprule
\textbf{Model}
& \textbf{MINC-2500}
& \textbf{NCT-CRC}
& \textbf{Places-LT}
& \textbf{Food-101} \\
\midrule

Haralick
& $22.9 \pm 0.2$
& $58.6 \pm 0.3$
& $0.6 \pm 0.1$
& $6.3 \pm 0.1$ \\

HOG
& $37.1 \pm 0.2$
& $58.7 \pm 0.2$
& $0.4 \pm 0.1$
& $8.5 \pm 0.2$ \\

WND-CHARM
& $42.8 \pm 0.1$
& $80.3 \pm 0.5$
& $8.4 \pm 0.2$
& $24.1 \pm 0.2$ \\

Fisher Vector
& $51.0 \pm 0.6$
& $76.8 \pm 0.6$
& $11.2 \pm 0.3$
& $34.9 \pm 0.2$ \\

\midrule

PRISM-HD
& $55.4 \pm 0.4$
& $87.2 \pm 0.4$
& $14.1 \pm 0.2$
& $36.1 \pm 0.2$ \\

\makecell[l]{PRISM-HD $\oplus$\\ Fisher Vector}
& $60.5 \pm 0.6$
& $89.4 \pm 0.3$
& $17.2 \pm 0.3$
& $48.1 \pm 0.2$ \\

PRISM-HD$_{+t}$
& $63.7 \pm 0.4$
& $88.3 \pm 0.7$
& $22.8 \pm 0.7$
& $58.4 \pm 0.2$ \\

\makecell[l]{\textbf{PRISM-HD$_{+t}$ $\oplus$}\\
\textbf{Fisher Vector}}
& $\mathbf{66.3 \pm 0.3}$
& $\mathbf{90.0 \pm 0.3}$
& $\mathbf{23.0 \pm 0.4}$
& $\mathbf{62.8 \pm 0.2}$ \\

\bottomrule
\end{tabular}

\end{table}

Two distinct patterns emerge from the comparison between PRISM-HD and Fisher Vector (FV). Although the artistic domain motivated our initial analysis, FV outperforms PRISM-HD on two of the three artistic datasets. One possible explanation is the different granularity of the two representations: FV explicitly aggregates local descriptors, whereas the current PRISM-HD instantiation relies primarily on global summaries of its derived views.

This pattern reverses on the non-artistic datasets, where PRISM-HD outperforms FV in all four cases. Readers familiar with Food-101 may notice that our FV result is slightly below previously published performance. We were unable to reproduce the reported values under our experimental protocol; nevertheless, the same qualitative comparison between PRISM-HD and FV is observed across the remaining non-artistic datasets.

More importantly, the relative ranking of PRISM-HD and FV does not determine the outcome of their combination. Across all seven datasets, their fixed equal-weight fusion outperforms either representation used in isolation. This consistent gain indicates that the two methods provide complementary predictive signals despite their varying standalone performance across domains. In other words, PRISM-HD and FV appear to emphasize different aspects of the visual signal, so their combination remains beneficial regardless of which representation is stronger on a particular dataset.

The final rows extend this fusion by incorporating the externally informed textual view. Adding text generally provides a further improvement over the purely visual PRISM-HD--FV combination, indicating that linguistic information can contribute an additional complementary signal. Importantly, however, the main fusion gain is already present without the textual channel, showing that the observed complementarity does not depend on externally pretrained linguistic information.

\subsection{Frozen backbones analyses}

A second evaluation allows the use of externally pretrained representations. Table~\ref{tab:frozen_results} compares PRISM-HD and its extensions with frozen ResNet-50, ViT-B/16, and CLIP ViT-B/32. Relative to conventional ImageNet-pretrained backbones, PRISM-HD remains highly competitive on the artistic datasets, outperforming both ResNet-50 and ViT-B/16 when combined with Fisher Vector and the textual channel. CLIP provides a substantially stronger reference, however, outperforming PRISM-HD on OilPainting and Multitask125, while PRISM-HD remains stronger on Rijksdraw.

A similar pattern is observed outside the artistic domain. CLIP considerably strengthens the frozen baseline, particularly on MINC-2500, Places-LT, and Food-101, while ViT-B/16 remains strongest on NCT-CRC. These results reinforce that PRISM-HD should not be interpreted as a replacement for large pretrained representations. Rather, the more relevant question is whether its structurally constructed views provide predictive information that remains complementary to such learned representations.

\begin{table}
\centering
\caption{
Balanced accuracy (\%) of frozen pretrained representations.
CLIP ViT-B/32 is evaluated with a higher-capacity classification head
(2048--1024--1024) and is included as a strong pretrained reference.
}\label{tab:frozen_results}

\small
\renewcommand{\arraystretch}{1.08}

\begin{tabular}{lccc}
\toprule
\textbf{Model}
& \textbf{OilPainting}
& \textbf{Multitask125}
& \textbf{Rijksdraw} \\
\midrule

ResNet-50 (frozen)
& $48.5 \pm 1.0$
& $35.7 \pm 0.9$
& $48.0 \pm 0.9$ \\

ViT-B/16 (frozen)
& $50.5 \pm 0.7$
& $36.1 \pm 2.0$
& $47.8 \pm 0.8$ \\

CLIP ViT-B/32
& $\mathbf{67.35 \pm 1.03}$
& $\mathbf{59.14 \pm 1.07}$
& $58.21 \pm 0.64$ \\

\makecell[l]{PRISM-HD$_{+t}$ $\oplus$\\ Fisher Vector}
& $54.3 \pm 0.8$
& $49.4 \pm 0.9$
& $\mathbf{61.6 \pm 0.6}$ \\

\bottomrule
\end{tabular}

\vspace{0.7em}

\begin{tabular}{lcccc}
\toprule
\textbf{Model}
& \textbf{MINC-2500}
& \textbf{NCT-CRC}
& \textbf{Places-LT}
& \textbf{Food-101} \\
\midrule

ResNet-50 (frozen)
& $67.6 \pm 0.4$
& $87.7 \pm 0.2$
& $30.6 \pm 0.2$
& $61.4 \pm 0.2$ \\

ViT-B/16 (frozen)
& $68.4 \pm 0.4$
& $\mathbf{91.9 \pm 0.3}$
& $33.2 \pm 0.2$
& $73.8 \pm 0.1$ \\

CLIP ViT-B/32
& $\mathbf{71.8 \pm 0.2}$
& $89.7 \pm 0.6$
& $\mathbf{43.3 \pm 0.4}$
& $\mathbf{88.4 \pm 0.1}$ \\

\makecell[l]{PRISM-HD$_{+t}$ $\oplus$\\ Fisher Vector}
& $66.3 \pm 0.3$
& $90.0 \pm 0.3$
& $23.0 \pm 0.4$
& $62.8 \pm 0.2$ \\

\bottomrule
\end{tabular}

\end{table}
\subsection{Backbone Fusion}

Although, as expected, even heterogeneous hand-crafted representations cannot consistently match the performance of fine-tuned deep models, this comparison raises a second and more relevant question: whether PRISM-HD captures predictive information that is complementary to the representations learned by deep networks. Rather than treating PRISM-HD and deep learning as competing approaches, we therefore evaluate whether combining them improves upon the corresponding fine-tuned backbone.

Table~\ref{deepcomp} summarizes the gains obtained through fixed equal-weight posterior fusion. Statistical significance was assessed using a paired permutation test on balanced accuracy, pairing the backbone and fused model under the same data splits and random seeds. Despite using neither a complex fusion strategy nor dataset-specific tuning, the fusion yields significant improvements ($p<0.05$) for both ResNet-50 and ViT-B/16 on the datasets where the largest gains are observed. On the remaining datasets, the gains are smaller and do not reach statistical significance; importantly, no statistically significant degradation is observed. These results indicate that PRISM-HD provides predictive signals complementary to those learned by the deep backbones, even under a simple parameter-free fusion rule.

\begin{table}[!htbp]
\centering
\caption{
Complementarity with fine-tuned deep models in balanced accuracy (\%).
$+P$ denotes fixed equal-weight posterior fusion with
PRISM-HD$_{+t}\oplus$ Fisher Vector. $\Delta$ denotes the change in mean
balanced accuracy relative to the corresponding fine-tuned backbone.
}
\label{deepcomp}

\small
\renewcommand{\arraystretch}{1.08}

\begin{tabular}{lccc|ccc}
\toprule
&
\multicolumn{3}{c}{\textbf{ResNet-50}}
&
\multicolumn{3}{c}{\textbf{ViT-B/16}} \\
\cmidrule(lr){2-4}
\cmidrule(lr){5-7}

\textbf{Dataset}
& \textbf{Base}
& \textbf{+P}
& $\boldsymbol{\Delta}$
& \textbf{Base}
& \textbf{+P}
& $\boldsymbol{\Delta}$ \\
\midrule

OilPainting
& $62.9 \pm 0.4$
& $65.6 \pm 1.4$
& $\mathbf{+2.7}$
& $61.8 \pm 0.4$
& $63.9 \pm 0.5$
& $\mathbf{+2.1}$ \\

Multitask125
& $50.1 \pm 0.4$
& $55.0 \pm 1.2$
& $\mathbf{+4.9}$
& $47.6 \pm 0.6$
& $56.1 \pm 1.1$
& $\mathbf{+8.5}$ \\

Rijksdraw
& $59.5 \pm 0.2$
& $62.6 \pm 0.4$
& $\mathbf{+3.1}$
& $57.8 \pm 0.6$
& $63.9 \pm 0.7$
& $\mathbf{+6.1}$ \\

\bottomrule
\end{tabular}

\vspace{0.7em}

\begin{tabular}{lccc|ccc}
\toprule
&
\multicolumn{3}{c}{\textbf{ResNet-50}}
&
\multicolumn{3}{c}{\textbf{ViT-B/16}} \\
\cmidrule(lr){2-4}
\cmidrule(lr){5-7}

\textbf{Dataset}
& \textbf{Base}
& \textbf{+P}
& $\boldsymbol{\Delta}$
& \textbf{Base}
& \textbf{+P}
& $\boldsymbol{\Delta}$ \\
\midrule

MINC-2500
& $81.7 \pm 0.3$
& $82.1 \pm 0.4$
& $\mathbf{+0.4}$
& $81.2 \pm 0.5$
& $81.8 \pm 0.5$
& $\mathbf{+0.6}$ \\

NCT-CRC
& $94.5 \pm 0.3$
& $95.0 \pm 0.4$
& $\mathbf{+0.5}$
& $93.7 \pm 0.9$
& $95.2 \pm 0.5$
& $\mathbf{+1.5}$ \\

Places-LT
& $33.9 \pm 0.5$
& $35.0 \pm 0.9$
& $\mathbf{+1.1}$
& $33.8 \pm 0.3$
& $35.0 \pm 0.2$
& $\mathbf{+1.2}$ \\

Food-101
& $87.3 \pm 0.1$
& $88.2 \pm 0.1$
& $\mathbf{+0.9}$
& $86.5 \pm 1.3$
& $87.4 \pm 1.3$
& $\mathbf{+0.9}$ \\

\bottomrule
\end{tabular}

\end{table}

\begin{figure}[!htbp]
    \centering
    \includegraphics[width=0.8\linewidth]{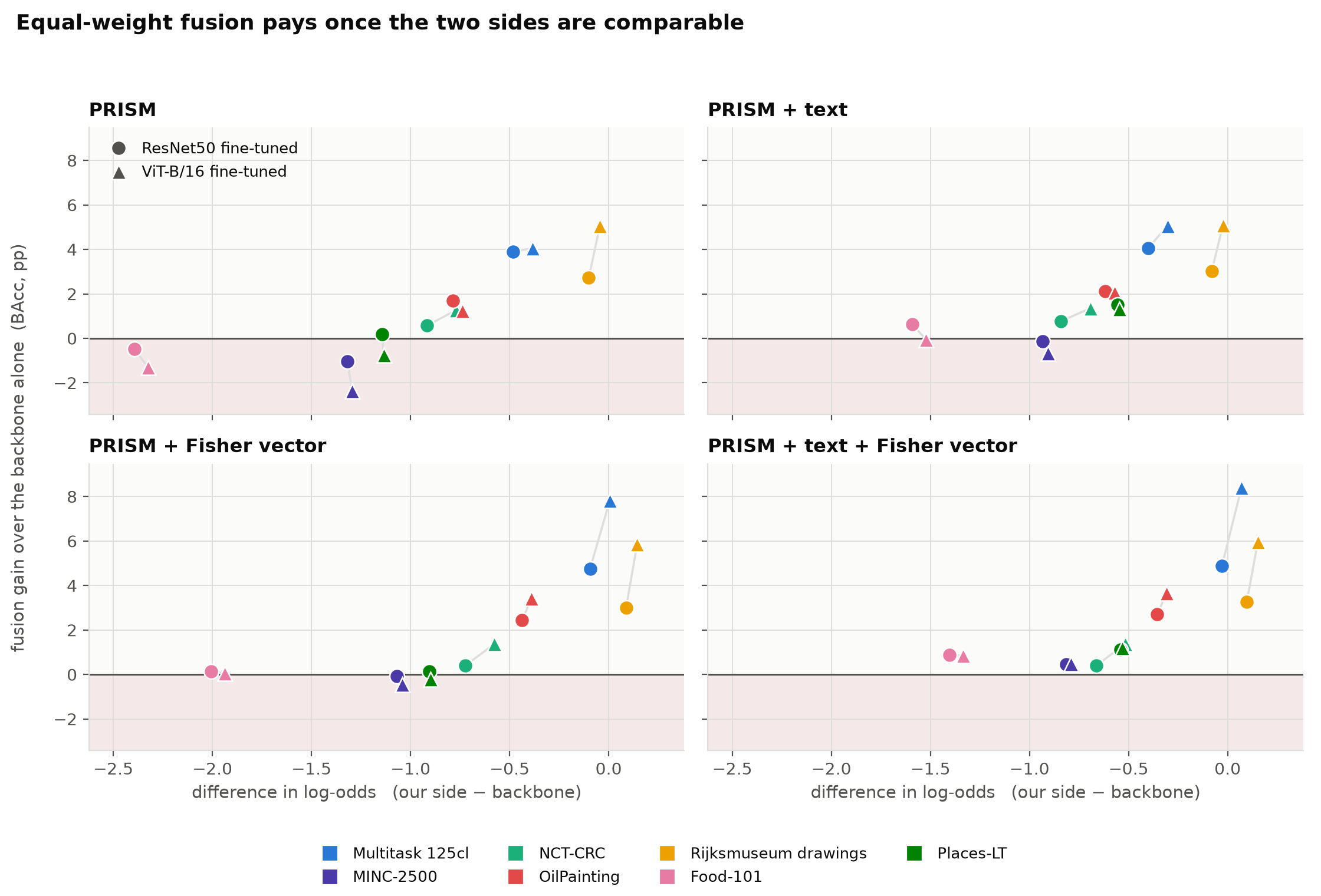}
    \caption{Fusion scenarios for different settings.}
    \label{fusion}
\end{figure}

A potential concern is that the observed fusion gains could be largely driven by the external knowledge introduced through the pretrained model used to construct the textual view. Figure~\ref{fusion} addresses this possibility by progressively adding PRISM-HD, Fisher Vector, and the textual channel to the deep backbones. PRISM-HD alone can occasionally reduce backbone performance, whereas incorporating Fisher Vector shifts the fusion toward consistently neutral or positive gains. This behavior already emerges before the textual channel is introduced, indicating that the main complementarity with deep representations does not depend on externally pretrained linguistic information.

The textual channel provides an additional but comparatively modest improvement to the final ensemble. Its contribution is also constrained by the fixed equal-weight fusion: when PRISM-HD remains substantially weaker than the backbone, as in the object- and scene-centric datasets, the gain provided by text is insufficient to close the performance gap. Conversely, on datasets where PRISM-HD is already competitive, the textual representation produces only a limited additional gain once combined with Fisher Vector and the backbone. Overall, the textual channel strengthens the ensemble, but does not account for the main complementarity observed between the hand-designed representations and deep models.

\section{Limitations}

PRISM has several limitations that should be considered when interpreting the results. First, the current set of structural maps is manually designed and is not intended to be exhaustive; different tasks may benefit from alternative representational domains. Second, the computational cost of the views is heterogeneous, and some transformations are substantially more expensive than conventional image descriptors. Third, the similarity graph is a dataset-contextual representation and therefore depends on the training set, rather than on the image alone. The textual view also differs from the other representations because it relies on an externally pretrained vision-language model and should therefore be interpreted as an externally informed view.

As discussed in Section~\ref{what}, the definition of what constitutes a view, and even what is considered the original image representation, is not unique. Different choices of structural maps may therefore lead to different sets of representations and potentially different conclusions. Likewise, the procedure used to summarize and combine the resulting views into a final predictive model introduces an additional modeling choice. Our experiments adopt specific feature extractors, classifiers, and fusion strategies, and the observed behavior should consequently be interpreted relative to these choices.

Finally, the reported gains provide evidence of predictive complementarity, but do not imply statistical independence between representations nor identify the precise information captured by each view. More generally, the empirical patterns observed in this work should not be interpreted as universal laws of single-source multi-view learning. Both the usefulness of a given view and the benefit of combining views are dataset- and model-dependent. PRISM-HD's standalone performance similarly varies across tasks, although its combination with Fisher Vector substantially reduces this variability across the evaluated domains.

\section{Conclusions}
\label{conclusion}

In this work, we investigated image classification through PRISM, a heterogeneous multi-view framework that transforms a common image into multiple structured representations, including graphs, sequences, spectral domains, and text. In the PRISM-HD instantiation, each view is summarized through hand-designed features and integrated through a branch-based MLP with view-specific encoders.

PRISM-HD was evaluated on three artistic and four non-artistic datasets. It remained competitive with conventional frozen backbones such as ResNet-50 and ViT-B/16 on the artistic datasets, although it generally remained below the stronger CLIP ViT-B/32 reference. Its standalone performance was more task-dependent outside the artistic domain, with stronger results on texture- and appearance-driven tasks than on object- and scene-centric recognition. However, this variability was substantially reduced when PRISM-HD was combined with Fisher Vector: their fixed, parameter-free fusion consistently improved over either shallow representation alone and provided additional gains when fused with fine-tuned deep backbones. Ablation analyses further showed that these improvements cannot be explained by feature dimensionality alone, and that the constructed views exhibit both redundancy and complementarity, with their usefulness depending partly on the downstream fusion architecture.

Overall, these results suggest that a single observation can support multiple predictively useful representations when its information is reorganized under different structural biases. Rather than replacing learned visual representations, PRISM provides a complementary approach to representation design and highlights the value of explicitly controlling how the same visual signal is exposed to downstream models.

\section*{Data and Code Availability}

All datasets used in this study are publicly available benchmark datasets and can be obtained from their respective official sources. The source code, experimental configurations, and scripts used to reproduce the experiments are publicly available at \href{https://github.com/Dudu-Campos/Paintings}{https://github.com/Dudu-Campos/Paintings}.

\section*{CrediT Authorship Contribution Statement}

\textbf{Eduardo Campos:} Software, Writing -- original draft, Writing -- review \& editing.  
\textbf{Willam Cardoso:} Software, Writing -- original draft, Writing -- review \& editing.   \\
\textbf{Diego Furtado:} Supervision, Writing -- review \& editing.   \\
\textbf{Solange Oliveira:} Supervision, Writing -- review \& editing.  \\
\textbf{Alneu Lopes:} Supervision, Writing -- review \& editing.

\section*{Declaration of Generative AI and AI-Assisted Technologies in the Manuscript Preparation Process}

During the preparation of this work, the authors used Claude Code (Anthropic) and ChatGPT (OpenAI) to assist with coding, figure generation, and language editing, including translation and paraphrasing of author-written text into English. All AI-assisted outputs were reviewed, verified, and edited by the authors, who take full responsibility for the content of the manuscript.

\section*{ACKNOWLEDGMENTS} \label{sec:ACKNOWLEDGMENTS}

This work was supported in part by the Coordination for the Improvement of Higher Education Personnel – Brazil (CAPES), Finance Code 001, by the Brazilian National Council for Scientific and Technological Development (CNPq) [Grants No. 134627/2025-3, 317002/2026-0, 309575/2021-4, 307184/2025-0, 303031/2023-9, and 101057/2024-5, 303588/2022-5], by the Academic Master's and Doctoral Program for Innovation (MAI/DAI) with a scholarship granted in partnership with BIRDIE.AI and administered by the Research Support Foundation (FUNAPE) [Scholarship No. 44519], by the Research Support Foundation of the State of Goiás (FAPEG) [Grant No. 23104.003594/2025-12,20/09835-1], and by the São Paulo Research Foundation (FAPESP) [Grant No. 2025/07160-0].
\clearpage
\clearpage

\bibliographystyle{cas-model2-names}
\bibliography{refs}

\end{document}